\documentclass[10pt,twocolumn,letterpaper]{article}

\usepackage[pagenumbers]{cvpr} 

\usepackage{graphicx}
\usepackage{amsmath, bm}
\usepackage{amssymb}
\usepackage{booktabs}
\usepackage{enumitem}

\usepackage[pagebackref,breaklinks,colorlinks]{hyperref}

\usepackage[capitalize]{cleveref}
\crefname{section}{Sec.}{Secs.}
\Crefname{section}{Section}{Sections}
\Crefname{table}{Table}{Tables}
\crefname{table}{Tab.}{Tabs.}

\def\confName{CVPR}
\def\confYear{2023}

\begin{document}

\title{MoFusion: A Framework for Denoising-Diffusion-based Motion Synthesis} 

\newcommand{\apriori    }     {\textit{a~priori}}
\newcommand{\aposteriori}     {\textit{a~posteriori}}
\newcommand{\perse      }     {\textit{per~se}}
\newcommand{\naive      }     {{na\"{\i}ve}}
\newcommand{\Naive      }     {{Na\"{\i}ve}}
\newcommand{\norm}[1]{\left\lVert#1\right\rVert}

\author{Rishabh Dabral$^1$\hspace{1.8em} Muhammad Hamza Mughal$^{1,2}$\hspace{1.8em}
Vladislav Golyanik$^1$\hspace{1.8em} Christian Theobalt$^{1}$\vspace{10pt}\\
$^1$Max Planck Institute for Informatics, SIC 
\hspace{6em}
$^2$Saarland University\\
}

\twocolumn[{ 
\renewcommand\twocolumn[1][]{#1} 
\maketitle 
\begin{center} 
    \vspace{-15pt} 
    \includegraphics[width=\textwidth ]{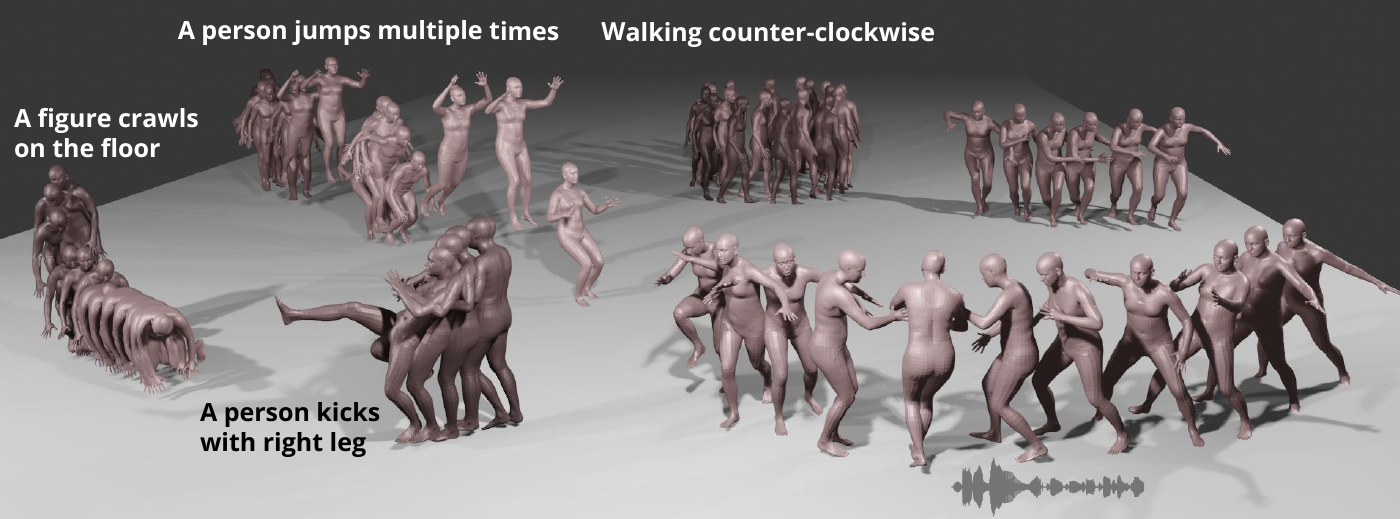} 
    \captionof{figure}{\textbf{Our MoFusion approach  synthesises long sequences of human motions in 3D from textual and audio inputs (\textit{e.g.,} by providing music samples)}. 
    Our model has significantly improved  generalisability and realism, and can be conditioned on modalities like text and audio.
    The resulting dance movements match the rhythm of the conditioning music, even if it is outside the training distribution. 
    }
    \label{fig:teaser} 
\end{center} 
}] 

\begin{abstract}
   Conventional methods for human motion synthesis have either been deterministic or have had to struggle with the trade-off between motion diversity vs~motion quality. 
   In response to these limitations, we introduce \textbf{MoFusion}, \textit{i.e.,} a new denoising-diffusion-based framework for high-quality conditional human motion synthesis that can synthesise long, temporally plausible, and semantically accurate motions based on a range of conditioning contexts (such as music and text). 
   We also present ways to introduce well-known kinematic losses for motion plausibility within the motion-diffusion framework through our scheduled weighting strategy. 
   The learned latent space can be used for several interactive motion-editing applications like in-betweening, seed-conditioning, and text-based editing, thus, providing crucial abilities for virtual-character animation and robotics.
   Through comprehensive quantitative evaluations and a perceptual user study, we demonstrate the effectiveness of MoFusion compared to the state of the art on established benchmarks in the literature. 
   We urge the reader to watch our supplementary video at \url{https://vcai.mpi-inf.mpg.de/projects/MoFusion/}.
\end{abstract}

\section{Introduction}\label{sec:intro} 
3D human motion synthesis 
is an important generative computer vision problem that often arises in robotics, virtual character animation and video games and movie production (\textit{e.g.,} for crowd dynamics simulation). 
It saw impressive progress over the last years; several works recently tackled it with reinforcement learning \cite{amp, bailando, nsm}, deep generative models \cite{moglow, rempe2021humor, temos, actor} or using deterministic approaches~\cite{aichoreo, ghosh, martinezetal}.
Despite the progress, multiple open challenges remain, such as improving motion variability, enabling higher motion realism and enhancing synthesis fidelity under  user-specified conditioning. 
Under \textit{conditioning}, we understand influencing the model outputs according to a control signal (\textit{e.g.,} ``walking counter-clockwise''). 
\par 
The key goal of conditional human motion synthesis 
is to generate motions that semantically agree with the conditioning while exhibiting diversity for the same conditioning signal. 
To facilitate the same, the recent state-of-the-art approaches have widely adopted generative techniques like conditional variational auto-encoders (CVAE)~\cite{action2motion, temos, actor, motionvae}, normalizing flows~\cite{moglow, aliakbarian2022flag}, as well as GANs~\cite{dual_motion_gan, inpainting_gan}. 
Naturally, each of them has strengths and limitations. 
GAN-based synthesis methods suffer from mode-collapse, thus resulting in insufficient fidelity of synthesis, especially for less common input conditioning. 
On the other hand, methods using CVAEs and normalizing flows typically have to deal with the trade-off between synthesis quality and the richness of the latent space (\ie, diversity)~\cite{aliakbarian2022flag, rempe2021humor}. 
\par 
The seminal works of Sohl-Dickstein \etal~\cite{ddpm} and Ho~\etal~\cite{hosalimans} recently demonstrated the ability of Denoising Diffusion Probabilistic Models (DDPM) to learn the underlying data distribution while also allowing for diverse sampling. 
Recent works~\cite{dalle2, stable_diffusion, glide} exhibited remarkable capabilities in the \textit{conditional} synthesis of images and audio with high-frequency details while also allowing interactive applications like editing and inpainting. 
\textit{
However, it has remained unclear how DDPM could be trained for such a problem with the temporal component as human motion synthesis.  
} 
\par 
Motivated by the recent advances in diffusion models, we propose MoFusion, \textit{i.e.,} a new approach for human motion synthesis with DDPM. 
This paper shows that diffusion models are highly effective for this task; see  Fig.~\ref{fig:teaser} for an overview. 
Our proposal includes a lightweight 1D U-Net network for reverse diffusion to reduce the rather long inference times. 
Furthermore, we demonstrate how domain-inspired kinematic losses can be introduced to diffusion framework during training, thanks to our time-varying weight schedule, which is our primary contribution. 
The result is a new versatile framework for human motion synthesis that produces diverse, temporally and kinematically plausible, and semantically accurate results. 
\par
We analyse DDPM for motion synthesis on two relevant sub-tasks: \textit{music-conditioned choreography generation} and \textit{text-conditioned motion synthesis.} 
While most existing choreography generation methods
produce repetitive (loopy) motions, and text-to-motion synthesis methods struggle with left-right disambiguation, directional awareness and kinematic implausibility, 
we show that MoFusion barely suffers from these limitations. 
Finally, formulating motion synthesis in a diffusion framework also affords us the ability to perform interactive editing of the synthesised motion. 
To that end, we discuss the applications of a pre-trained MoFusion, like motion forecasting and in-betweening, which are important applications for virtual character animation. 
We show improvements in both sub-tasks through quantitative evaluations on AIST++~\cite{aichoreo} and HumanML3D~\cite{humanml3d} datasets as well as a user study. 
In summary, our core \textbf{technical contributions} are as follows: 
\begin{itemize} 
\setlength{\parskip}{0pt} 
    \item The first method for conditional 3D human motion synthesis using denoising diffusion models. 
    Thanks to the proposed time-varying weight schedule, we incorporate several kinematic losses that make the synthesised outputs temporally plausible and semantically accurate with the conditioning signal. 
    \item Model conditioning on various signals, \textit{i.e.,} music and text, which is reflected in our framework's architecture. 
    For music-to-choreography generation, our results generalise well to new music and do not suffer from degenerate repetitiveness. 
\end{itemize} 

\section{Related Works}
\label{sec:related_works} 
We discuss the relevant literature from two vantage points,  \textit{i.e.,} prior methods for human motion synthesis and  literature on diffusion models. 
\subsection{Conditional Human Motion Synthesis} 

Traditionally, the problem of Human Motion Synthesis has been approached either by statistical modelling~\cite{stat_1,stat_2} or sequence modelling techniques~\cite{lstmmotion, martinezetal}. 
Both approaches employed an initial seed sequence corresponding to a starting pose or past motion, which helps guide future motion prediction.
However, synthesising motion sequences from scratch proves to be a harder task, where synthesis is guided by a conditioning mechanism. 
\par
A common approach in conditioned human motion synthesis is to guide the motion generation by using of class descriptions corresponding to actions~\cite{actor, action2motion}. 
These approaches typically employ generative models like conditional VAEs~\cite{vae} and learn a latent representation for motion based on action conditioning.
Among such methods, Action2Motion~\cite{action2motion} uses a frame-level motion representation with temporal VAEs, while ACTOR~\cite{actor} improves results using a sequence-level motion representation with transformer-VAEs to synthesise motions based on action input. 
However, action conditioning does not provide a rich description of the target motion.

\noindent \textbf{Text-Conditioned Motion Synthesis:} 
The methods discussed above were followed by text-conditioned motion synthesis, developed on textually-annotated motion datasets like KIT~\cite{kit}, BABEL~\cite{babel} and HumanML3D~\cite{humanml3d}. 
Such methods typically learn a shared latent space upon which both text and motion signals are projected~\cite{motionclip, ghosh}. 
Lin \etal~\cite{linetal} use an LSTM encoder and a GRU decoder to predict future pose sequences. %
Ahuja \etal~\cite{jl2p} and Ghosh \etal~\cite{ghosh} focus on creating a joint language and pose representation to synthesise the motions autoregressively. 
TEMOS~\cite{temos} builds upon the ideas by Ghosh \etal~\cite{ghosh} and ACTOR~\cite{actor} by using a transformer-VAE-based generative model with conditioning from a pre-trained language model. 
Finally, Guo \etal~\cite{humanml3d} use a temporal VAE to synthesise motions by extracting text-based features and then auto-regressively generating motion sequences. 
\par
\noindent \textbf{Dance-Conditioned Motion Synthesis:}
Besides text, audio has also been applied to guide human motion synthesis. 
Speech is used to learn gesture animations to mimic face, hand and body movements while speaking \cite{habibie, moglow}.
Similarly, dance music has also been used extensively to synthesise motions. 
Various works~\cite{audbodydynamics, selfsupaud, tempmusicbody, ginosar, deepphase} for music-conditioned motion synthesis tackle this problem by predicting motion from audio without seed motion.
However, they converge to a mean pose, as dance typically consists of repetitive poses. 
Li \etal~\cite{li_dance} address this problem by providing an initial pose and the audio as input to a transformer-based architecture.
DanceNet~\cite{dancenet} proposes an autoregressive generative model, while Dance Revolution~\cite{dancerev} uses a curriculum learning approach and a seq2seq architecture to synthesise dance motion. 
AI Choreographer~\cite{aichoreo} also approaches this problem by providing seed motion along with music to a cross-modal transformer for future dance motion prediction. 
Zhou \etal~\cite{zhouaristidou} enhance dance motions with music-to-dance alignment, and Aristidou \etal~\cite{aristidou} enforce a global structure of the dance theme over the motion synthesis pipeline.
The recent Bailando method~\cite{bailando} achieves impressive results in music-to-dance generation through a two-stage generation process. 
Their method learns to encode dance features into a codebook using a VQ-VAE~\cite{vqvae} and then employs GPT~\cite{gpt} to predict a future pose code sequence given input music and starting seed pose. 
Finally, the pose code sequence is converted into a dance sequence via the learned codebook and CNN decoder. 
Unlike Bailando~\cite{bailando}, our motion generation  does not require multiple stages during inference.
\par
Most of the existing methods depend on seed motion as input and usually produce repetitive dance choreography, and we differ from previous music-conditioned choreography generation methods by producing non-repetitive choreographies while also not requiring any seed motion sequence.
Moreover, earlier methods~\cite{li_dance, bailando, aichoreo} use handcrafted music features (such as beats, chroma and onset strength) along with MFCC representation of audio signals for predicting music-aligned dance sequences. 
In contrast, our method learns to predict dance sequences on raw Mel spectrograms without auxiliary features  like beats. 

\subsection{Diffusion Models} 
Diffusion models~\cite{dickstein} have shown great promise in terms of generative modelling by showing outstanding results in synthesis applications ranging from image generation~\cite{dalle2,imagen, hosalimans, stable_diffusion}, speech synthesis~\cite{diffwave, gradtts}, to point-cloud generation~\cite{luo2021diffusion}. 
The seminal work of Sohl-Dickstein \etal~\cite{dickstein}
gradually diffuses Gaussian noise into a training sample and trains a neural network to reverse-diffuse the noise. 
Ho \etal~\cite{ddpm}  apply the same modelling technique in DDPM to achieve high-quality image synthesis, and Song \etal~\cite{ddim} improve the efficiency of the generative process by introducing faster sampling in the reverse process. 
\par  
These models have been applied for various computer vision tasks like text-to-image generation. 
Paradigms like classifier guidance~\cite{dhariwaletal} and classifier-free guidance~\cite{hosalimans} for the diffusion process have been introduced to improve image synthesis quality. 
CLIP-based guidance strategies are also used by GLIDE~\cite{glide}. 
Ramesh \etal~\cite{dalle2} also utilise text-image embeddings by CLIP and a diffusion decoder to achieve high-quality image synthesis. 
Other than text-to-image generation tasks, diffusion models have also been popular in other vision applications~\cite{diffusionsurvey}. 
Besides image generation, diffusion models have also been applied to synthesise audio. 
Grad-TTS~\cite{gradtts} and DiffWave~\cite{diffwave} apply the diffusion paradigm to text-to-speech synthesis. 
Work by Luo \etal~\cite{luo2021diffusion} also uses diffusion models for 3D point cloud generation tasks. 
\par 
We note the presence of three concurrent works (published on arXiv at the time of submission) that are similar to our approach~\cite{mdm, motiondiffuse, flame}. 
However, all three methods differ in their network design and loss functions.
While FLAME~\cite{flame}, Tevet \etal~\cite{mdm} and Zhang \etal~\cite{motiondiffuse} use a transformer network, we instead choose a 1D U-Net with cross-modal transformers to learn the denoising function. 
We also train our network differently using a time-varying weighting schedule on the kinematic losses.
Finally, all concurrent works use diffusion models to synthesise motions conditioned on text and action.
On the other hand, we focus not only on text-driven motion synthesis but also on \textit{dance choreography generation using raw music}. 
\par
\begin{figure}[!t]
    \vspace{-15pt} 
    \includegraphics[width=\linewidth ]{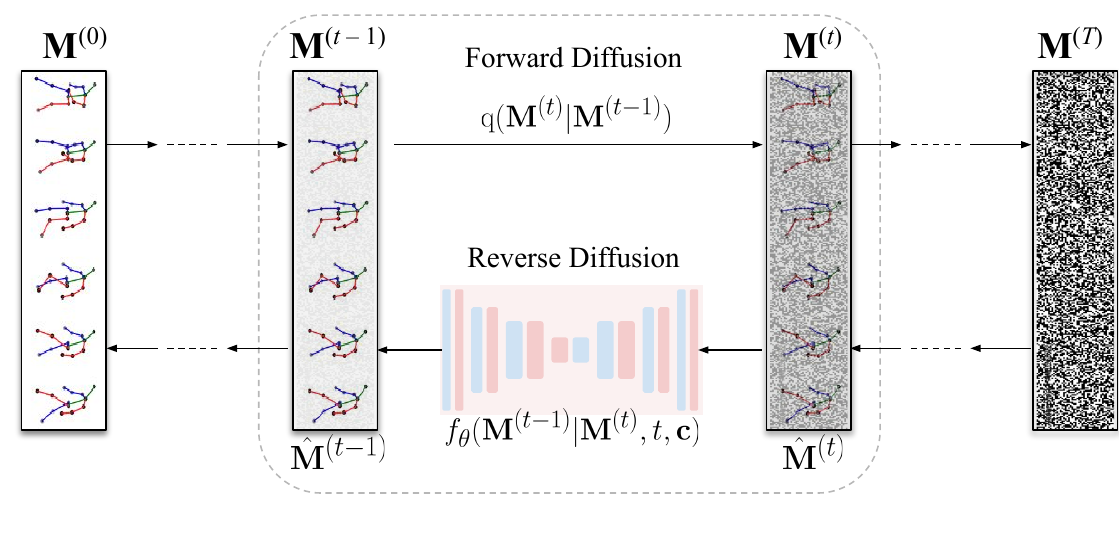} 
    \captionof{figure}{\textbf{An illustration of our diffusion for 3D motion synthesis.} During forward diffusion, we iteratively add Gaussian noise $q(\mathbf{M}^{(t)} | \mathbf{M}^{(t-1)}) = \mathcal{N}(\mathbf{M}^{(t)}|(1-\beta_t)\mathbf{M}^{(t-1)}, \beta_t\bm{I})$ to initial motion at $t{=}0$. A neural network $f_{\theta}(\cdot,\cdot)$ is trained to denoise the noisy motion $\mathbf{M}^{(t)}$ at time $t$ based on the conditioning signal $\mathbf{c}$. 
    } 
    \label{fig:overall_schema} 
\end{figure} 

\section{Method}
\label{sec:method} 
\begin{figure*}[!ht]
    \vspace{-15pt} 
    \includegraphics[width=\textwidth ]{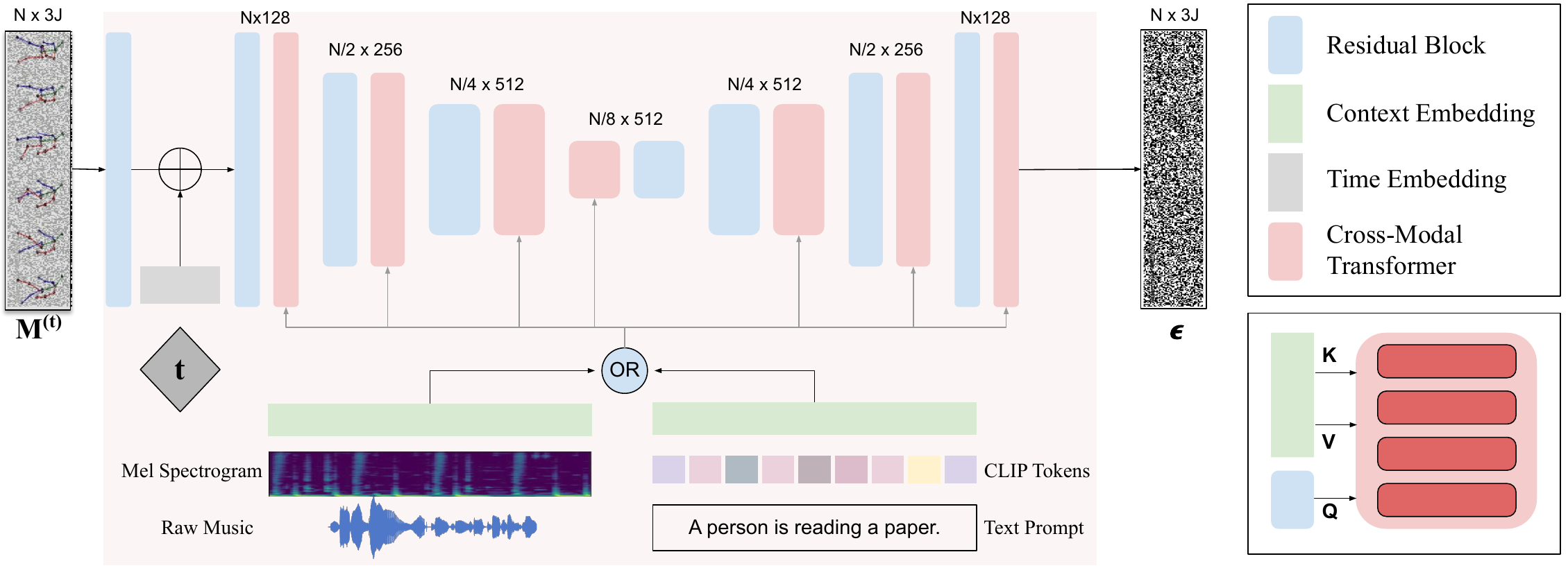} 
    \captionof{figure}{\textbf{Illustration of the 1D U-Net architecture with cross-modal transformer blocks with multi-head attention (bottom right)}. 
    The network's input is a noisy motion sample at timestep $t$, and the output is an estimate of the noise $\bm{\epsilon}$. 
    Additionally, it can be conditioned on either music or text prompts. 
    In both cases, we learn a projection function to map the conditioning features to 1D U-Net features.
    }
    \label{fig:architecture}  
\end{figure*}
Given a conditioning signal, $\mathbf{c} \in \mathbb{R}^{k\times d}$, our goal is to synthesize human motion $\mathbf{M}^{(0)}= \{\mathbf{m}_1, \mathbf{m}_2, \dots, \mathbf{m}_N\}$. 
The pose at each timestep $i$ is parameterised as $\mathbf{m}_i \in \mathbb{R}^{3J}$, which includes the  root-relative 3D coordinates of each of the $J$ joints and the  camera-relative translation of the root joint.
This representation is flexible and one could, if desired, train for joint angles instead (see supplementary materials). 
The conditioning signal $\bm{c}$, could either be an audio clip or a text prompt. 
It is represented as a $d$-dimensional embedding of $k$ Mel spectrogram features (for audio) or word tokens (for text). 

In the following, we first discuss the basics of denoising diffusion models (Sec.~\ref{ssec:diffusion_motion_synthesis}). 
Next, we discuss how our kinematic losses can be incorporated within the diffusion framework (Sec.~\ref{ssec:training}). 
Finally, the neural architecture design and the modifications required for task-specific conditioning are introduced (Sec.~\ref{ssec:MoFusion}). 
\subsection{Diffusion for Motion  Synthesis}\label{ssec:diffusion_motion_synthesis} 
%
%
The motion generation task is formulated as a reverse diffusion process that requires sampling a random noise vector, $\mathbf{z} \in \mathbb{R}^{N\times3J}$, from a noise distribution 
to generate a meaningful motion sequence (see Fig.~\ref{fig:overall_schema}). 
While training, the forward diffusion process requires successively corrupting motion sequence $\mathbf{M}^{(0)}$ by adding Gaussian noise to a motion sequence for $T$ timesteps in a Markovian fashion.
This results in the conversion of a meaningful motion sequence $\mathbf{M}^{(0)}$ in the training set into a noise distribution $\mathbf{M}^{(T)}$:
\begin{equation}
   q \big(\mathbf{M}^{(1:T)} | \mathbf{M}^{(0)} \big) = \prod_{t=1}^{t=T} q\big(\mathbf{M}^{(t)} | \mathbf{M}^{(t-1)}\big), 
\end{equation}
where $q(\mathbf{M}^{(t)} | \mathbf{M}^{(t-1)}) = \mathcal{N}(\mathbf{M}^{(t)}|(1-\beta_t)\mathbf{M}^{(t-1)}, \beta_t\bm{I})$ is the Markov diffusion kernel that adds Gaussian noise to the motion at time step $t$, and $\beta_t$ is a hyperparameter that controls the rate of diffusion. 
In practice, there exists a re-parameterisation trick that allows closed-form sampling at any timestep $t$: 
\begin{equation} \label{eq:forward_diffusion}
    \mathbf{M}^{(t)} = \sqrt{\bar{\alpha_t}}\,\mathbf{M}^{(0)} + \sqrt{1 - \bar{\alpha_t}}\,\bm{\epsilon}, 
\end{equation}
wherein $\bm{\epsilon}$ is the random noise matrix and $\bar{\alpha_t} = \prod_{s=0}^{t} (1 - \beta_s)$. 
With sufficiently large $T$, one can assume  $\mathbf{M}^{(T)} \approx \mathbf{z}$.
\par
To generate a motion sequence from a random noise matrix $\mathbf{z}$, we need to iteratively reverse-diffuse $\mathbf{z}$ for $T$ timesteps.
The reverse-diffusion is formulated as~\cite{dickstein}:
\begin{equation}
    p\big(\mathbf{M}^{(0:T)}\big) = p\big(\mathbf{M}^{(T)}\big) \prod_{t=1}^T p\big(\mathbf{M}^{(t-1)} | \mathbf{M}^{(t)}\big). 
\end{equation}
The reverse transition probability $p(\mathbf{M}^{(t-1)} |\mathbf{M}^{(t)})$ is approximated using a neural network that learns the function $f_{\theta}(\mathbf{M}^{(t-1)} | \mathbf{M}^{(t)}, t)$.
While several variations of $f_\theta(\cdot,\cdot)$ exist, we follow~\cite{hosalimans} and train the network to predict the original noise $\bm{\epsilon}$.
For the conditional synthesis setting, the network is additionally subjected to the conditioning signal $\bm{c}$ as $f_\theta(\mathbf{M}^{(t-1)} | \mathbf{M}^{(t)}, t, \bm{c})$.

\subsection{Training Objectives} 
\label{ssec:training}
We now discuss how kinematic loss terms inspired by domain knowledge can be introduced within the diffusion framework.
The overall loss for training MoFusion is a weighted sum of two broad loss types:
\begin{equation} 
    \mathcal{L}_t = \mathcal{L}_{da} +  \lambda_k^{(t)}\mathcal{L}_k. 
\end{equation} 
The primary data term, $\mathcal{L}_{da}$, is the commonly-used $L_2$ distance between the noise $\bm{\epsilon}$ used for forward diffusion \eqref{eq:forward_diffusion} and the estimated $f_{\theta}(\mathbf{M}^{(t)}, t, \mathbf{c})$. 
\par
While $\mathcal{L}_{da}$ is strong enough to approximate the underlying data distribution, the synthesised motions are not guaranteed to be physically and anatomically plausible.
Consequently, it allows for artefacts like motion jitter, illegal skeletons and foot-sliding.
Fortunately, human motion capture literature consists of several kinematic and physical constraints that can be used to regularize the synthesised motion~\cite{Zhou_2017_ICCV, Dabral2018, rempe2021humor, physcap}.
These kinematic loss functions are well established in the motion synthesis literature and have been consistently used to avoid synthesis artefacts.
However, since the denoising network is trained to estimate the \textit{noise} $\bm{\epsilon}$, it is not straightforward to apply such constraints.
One workaround is to apply the losses to the final reverse-diffused motion, which can be estimated using the re-parameterisation trick:
\begin{equation}
\mathbf{\hat{M}}^{(0)}= \frac{1}{\sqrt{\bar{\alpha_t}}}\mathbf{M}^{(t)} - \bigg(\sqrt{\frac{1}{\bar{\alpha}} - 1}\bigg)f_{\theta}\big(\mathbf{M}^{(t)}, t, c\big). 
\end{equation}
However, \naive{}ly using $\mathbf{\hat{M}}^{(0)}$ to approximate the reverse-diffusion outputs leads to unstable training because the generated motion is extremely noisy when $t$ is close to $T$. 
\par
Therefore, we introduce a time-varying weight schedule for $\mathcal{L}_k$ by varying the schedule as per $\lambda_k^{(t)} = \bar \alpha_t$. 
This ensures that the motions at $t{\approx}T$ receive an exponentially lower weight compared to $t{\approx}0$.
Within $\mathcal{L}_k = \mathcal{L}_s + \lambda_a\mathcal{L}_a + \lambda_m\mathcal{L}_m$, we include three loss terms:
First, we use the skeleton-consistency loss, $\mathcal{L}_s$, that ensures that the bone lengths in the synthesised motion remain consistent across time.
To achieve this, we minimize the temporal variance of the bone lengths, $l_n$:
\begin{equation}
    \mathcal{L}_s = \frac{\sum_n (l_n - \bar{l})^2}{n-1}, 
\end{equation}
where $\bar{l}$ is the vector of mean bone lengths.
Secondly, we use an anatomical constraint, $\mathcal{L}_a$, that penalizes left/right asymmetry of the bone lengths: $\mathcal{L}_a = \vert \vert \operatorname{BL}(j_1, j_2) - \operatorname{BL}(\delta(j_1), \delta(j_2))\vert \vert$, where $\operatorname{BL}(\cdot ,\cdot)$ computes the bone-lengths between the input joints and $\delta(\cdot)$ provides the index of the corresponding symmetrically opposite joint.
When using joint angle representation instead of joint positions, it is possible to use joint-angle limit regularisations as in~\cite{hmr} instead of bone length constraints.
Finally, we again add ground-truth supervision on motion synthesis, this time with: 
\begin{equation} 
\mathcal{L}_m = \norm{\hat{\mathbf{M}}^{(0)} - \mathbf{M}^{(0)}}_2. 
\end{equation} 

It is worth noting that these kinematic loss terms are not exhaustive and there exist several other loss terms that can attend to different aspects of motion synthesis.
For example, it is possible to add the foot-sliding loss of~\cite{motionet}, or physics-based constraints of~\cite{physcap, neural_physcap, rempe2021humor}. 
Through our formulation, we demonstrate how such losses can be incorporated within the \textit{diffusion framework}. 

\subsection{The MoFusion Architecture}\label{ssec:MoFusion} 
Drawing inspiration from successful 1D-Convnet architectures for motion synthesis~\cite{quaternet} and pose estimation~\cite{videopose3d}, we use a 1D U-Net~\cite{unet} to approximate $f_{\theta}(\cdot,\cdot)$.
This is also consistent with several state-of-the-art diffusion-based image generation methods~\cite{dalle2, stable_diffusion, imagen} that use a U-Net architecture for the denoising network.
The fully-convolutional nature of the network allows us to train the network with motions of various lengths.
Fig.~\ref{fig:architecture} illustrates the schema of the network.
The network consists of three downsampling blocks that first successively reduce the feature length, $n$, from $N$ to $\lfloor N/8\rfloor$ before being upsampled using corresponding upsampling blocks. 
Each 1D residual block is followed by a cross-modal transformer that incorporates the conditioning context, $\mathbf{c}$, into the network.
The time-embedding is generated by passing the sinusoidal time embedding through a two-layer MLP.
For incorporating the context, we treat the intermediate residual motion features, $\mathbf{x} \in \mathbb{R}^{n\times d}$, to get the query vector while using the conditioning signal, $\mathbf{c} \in \mathbb{R}^{m\times d}$,  to compute the key and value vectors. 
Specifically, we first estimate
\begin{equation}
\mathbf{Q} = W_q \mathbf{x},\,\,\mathbf{K} = W_k \mathbf{c},\,\,\text{and}\,\,\mathbf{V} = W_v \mathbf{c},
\end{equation}
where $W_q, W_k,$ and $W_v$ are the Query, Key and Value matrices, respectively.
As in standard cross-attention~\cite{vaswani}, the relevance scores are first computed with the softmax, and then used to weigh the values $\mathbf{V}$:
\begin{equation}
    \operatorname{Attention}(\mathbf{Q}, \mathbf{K}, \mathbf{V}) = \operatorname{softmax}\bigg(\frac{\mathbf{Q}\mathbf{K}^{\prime}}{\sqrt{d}}\bigg)\mathbf{V}. 
\end{equation}
In the case of unconditional generation, the formulation switches to self-attention by also getting the key and value vectors $\mathbf{K}, \mathbf{V}$ from $\mathbf{x}$.
We now discuss the task-dependent processing of the conditioning input.\\

\noindent \textbf{Music-to-Dance Synthesis:}
For conditioning the network to music signals, we choose to represent them using the Mel spectrogram representation~\cite{gradtts, tacotron2}.
This is unlike several existing music-to-dance synthesis methods~\cite{aichoreo, bailando} that use MFCC features along with music-specific features like beats or tempograms. 
Thus, we leave it up to the context-embedding layer to learn an appropriate projection to the feature space of U-Net.
In theory, this also allows our method to be trained on other \textit{audio} (not necessarily music) conditioning such as speech. 
\par
To extract the Mel spectrograms, we re-sample audio signals to 16kHz and convert them to log-Mel spectrograms with $k{=}80$ Mel bands by using hop-length of $512$ and the minimum and maximum frequencies of $0$ and $8$ kHz, respectively. 
As a result, we obtain a conditioning signal $\mathbf{c} \in  \mathbb{R}^{(m \times k)}$, where $m{=}32$ for one second of the audio signal. 
We use a linear layer to project the input Mel spectrogram onto the context embedding $\mathbf{c}$. \\

\noindent \textbf{Text-to-Motion Synthesis:}
Text-conditioned Diffusion Models~\cite{dalle2, stable_diffusion} have recently shown impressive generation capabilities.
For synthesizing motion from textual descriptions, we use the pre-trained CLIP~\cite{clip} token embeddings.
We first retrieve the tokenised embedding for each word in the input prompt.
Next, these token embeddings are position-encoded and subjected to CLIP's transformer.
Finally, we project the token embeddings using an MLP that maps the transformer embedding onto $\mathbf{c}$.

\section{Experiments} 
\label{sec:experiments} 
We next evaluate the proposed MoFusion framework in two  scenarios, \textit{i.e.,} conditioned by audio and text. 
We first discuss music-to-choreography generation details  (Sec.~\ref{sec:music_conditioning}), followed by text-conditioned motion generation (Sec.~\ref{sec:text_conditioning}) and, finally, show  applications like seed-motion forecasting, editing and inbetweening in Sec.~\ref{sec:editing}. 
\subsection{Music-to-Dance Synthesis}
\label{sec:music_conditioning}
\noindent \textbf{Datasets:} We train MoFusion for music-conditioned dance synthesis on the AIST++ Dataset ~\cite{aichoreo}.
The dataset contains $1408$ unique dance motion sequences with 
lengths ranging from $7.4$ to $48.0$ seconds. 
There are ten different dance motion genres with multiple dance choreographies for each genre, which provides a rich diversity in terms of types of dance motions.
The data has been annotated using multi-view capture and we use the provided 3D motion sequences as the target motion and their corresponding music as our conditioning input. 
More importantly, we use the dataset split based on music choreography, which ensures that the validation/test set contains unheard music and, correspondingly, choreography vis-\`a-vis the training set. 

\noindent \textbf{Evaluation Metrics:}
We perform the quantitative evaluation for music-conditioned synthesis by using Frechet Inception Distance (FID) score, Diversity (Div), Beat Alignment Score (BAS) and Multi-Modality.
The FID score is evaluated following the method used in Siyao \textit{et al.}~\cite{bailando}. 
We measure and compare FID using a kinetic feature extractor~\cite{kin_fid}, which includes \textit{hand-crafted} features regarding velocity and acceleration in its feature representation. 
We use implementation by the \texttt{fairmotion} toolbox~\cite{fairmotion} to measure FID. 
To measure the diversity of generated motions, the diversity metric (Div) computes the average pairwise Euclidean distance of the kinetic features of the motions synthesised from audios in the test set.
We also measure Beat Alignment Score (BAS)~\cite{aichoreo}, which expresses the similarity between the kinematic and music beats.
Here, kinematic beats refer to the local minima of the kinetic velocity of a motion sequence showing beats as the ``stopping points'' during the motion. Moreover, music beats in the audio signal are extracted using Librosa toolbox~\cite{librosa}. 
The score is defined as the mean distance between every kinematic beat and its nearest music beat: 
\begin{equation} \label{eq:beatalignscore}
\small
	BAS = \frac{1}{|B^m|} \sum_{b^m \in B^m} \exp \bigg(-\frac{\min_{\forall b^{d} \in B^{d}} \parallel b^{d} - b^{m} \parallel^2}{2{\sigma}^2}\bigg), 
\end{equation}
where $b^{d}$ represents a kinematic beat with $B^d$ being a set of all kinematic beats and $b^{m}$ represents a music beat with $B^m$ being a set of all music beats. We follow \cite{bailando,aichoreo} and keep $\sigma=3$ in our experiments. 
Finally, we also measure multi-modality for our approach by calculating the average Euclidean distance between the kinetic features of $K{=}50$ generated motion sequences for the same music input. This expresses the multi-modality of the dance generation.
\par
\noindent \textbf{Quantitative Results:} 
The quantitative results are summarised in Table~\ref{tab:audio}. 
Our method improves upon the \textit{Diversity} scores of the state of the art and 
achieves a multi-modality score of \textbf{11.38}.
These results confirm the variability claims of DDM for motion synthesis.
In contrast, state-of-the-art methods like Bailando and AI Choreographer are deterministic and produce similar outputs given the same input music. 
Therefore, measuring multi-modality is not applicable to them.
In addition, we observe a better beat alignment score than Bailando~\cite{bailando} and the ground truth, showing that MoFusion learns better motion alignment with beats.
It is also worth noting that Bailando explicitly uses BAS in its reward formulation, whereas we do not.
Finally, we observe subpar performance on FID compared to~\cite{bailando, aichoreo}.
Upon visual inspection, we notice that both Bailando and AI-Choreographer produce \textit{repetitive}, loopy dance motions which are very similar to the ground truth (and the training set).
On the other hand, our diffusion-based model seldom produces repetitive or loopy motions and, therefore, differs significantly from the hand-crafted kinetic-feature profile used to compute the FID.

\begin{table}[!t]
\resizebox{\columnwidth}{!}{%
\begin{tabular}{lcccc}
\hline
                                        & \multicolumn{2}{c}{Quality}             & \multicolumn{2}{c}{Diversity} \\ \cline{2-5} 
Method                                  & BAS~$\uparrow$       & FID~$\downarrow$   & Div~$\uparrow$    & M.-Modality~$\uparrow$       \\ \hline
Ground Truth                            & 0.237                & 17.10              & 8.19              & n/a     \\ \hline
Li \etal~\cite{li_dance}                & 0.160                & 86.43              & 6.85              & n/a     \\
DanceNet~\cite{dancenet}                & 0.143                & 69.13              & 2.86              & n/a     \\
Dance Revolution~\cite{dancerev}        & 0.195                & 73.42              & 3.52              & n/a     \\
AI Choreographer~\cite{aichoreo}        & 0.221                & 35.35              & 5.94              & n/a     \\
Bailando~\cite{bailando}$^\dagger$      & 0.233                & \textbf{28.16}     & 7.83              & n/a     \\ \hline
MoFusion (\textit{Ours})                & \textbf{0.253}       & 50.31              & \textbf{9.09}     &\textbf{11.38}      \\ \hline
\end{tabular}
}
\caption{Comparison of our method with the previous  methods. We achieve state-of-the-art performance on beat alignment score as well as Diversity. ``$^\dagger$'': Unlike Bailando~\cite{bailando}, we do not explicitly train our method using BAS as a reward or a loss function.} 
\label{tab:audio}
\end{table}

\begin{figure}[h]
    \includegraphics[width=\linewidth ]{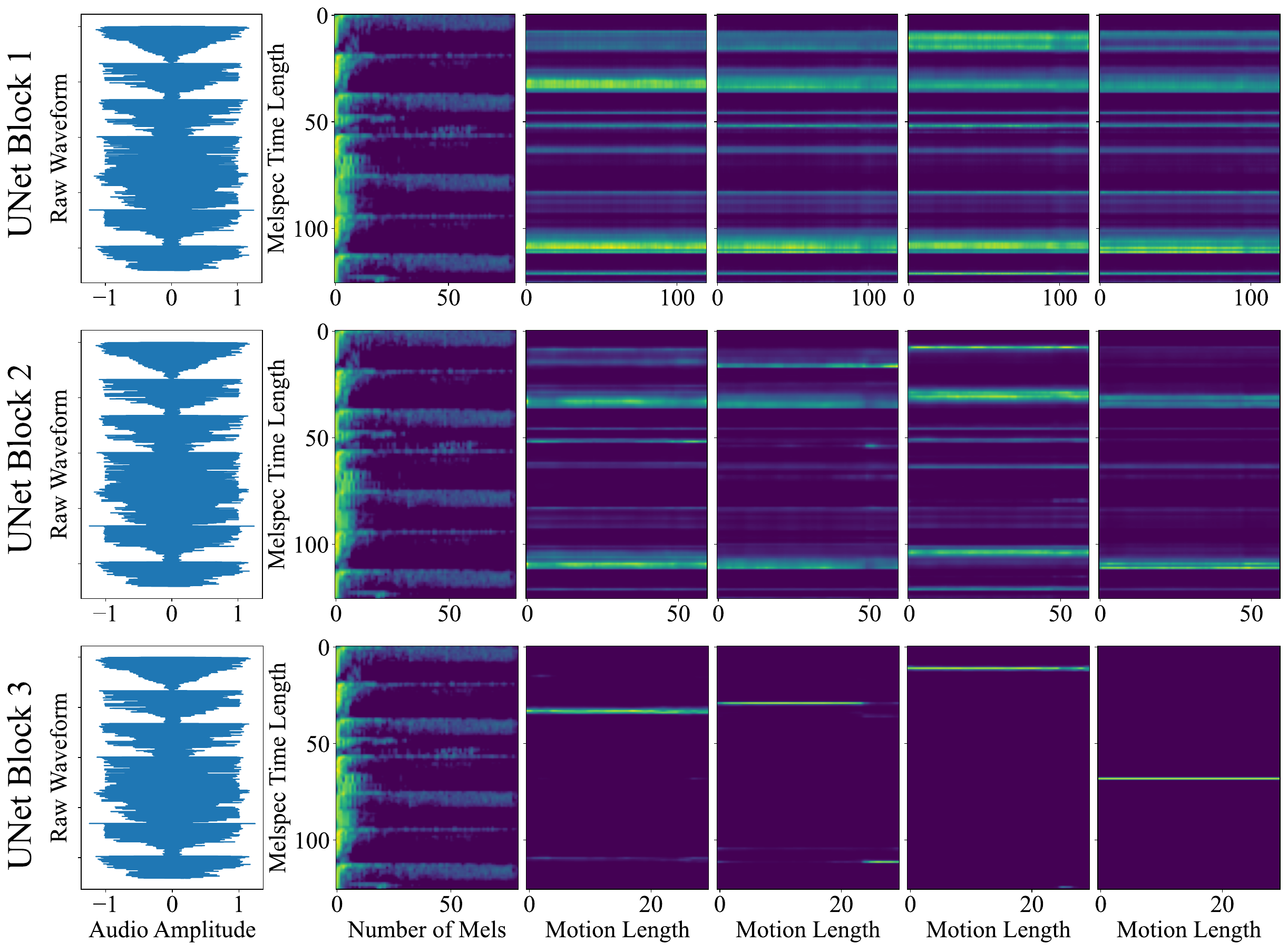}
    \captionof{figure}{Visualisation of cross-modal attention weights at different levels of the U-Net. 
    Notice the alignment of the attention weights to the specific beats in the audio.
    Also, while the shallower levels (top) have scattered attention, the attention heads at the bottleneck layer (bottom) degenerate to specific audio sections corresponding to the music beats.
    }
    \label{fig:attention_weights}  
\end{figure}

\noindent \textbf{Analysis:} Fig.~\ref{fig:attention_weights} depicts the cross-modal attention weights of the audio signal against the generated motion.
Interestingly, we observe that the transformer learns to associate high attention with the occurrence of \textit{beats} in the music. 
Here, the beats are not provided as input features, and beat  alignment is automatically learnt from the Mel spectrogram by the network. 
This is in contrast to methods that either explicitly use music-specific hand-crafted features~\cite{aichoreo} or train the network with a beat alignment loss~\cite{bailando}.
Upon qualitative inspection, we also notice that, unlike other methods, our synthesised choreography rarely repeats (see supplementary video). 
MoFusion manages to avoid this phenomenon since we do not require a seed motion input that can bias the network towards loopy motion. 

\subsection{Text-to-Motion Synthesis}
\label{sec:text_conditioning}
\noindent \textbf{Datasets:} 
For the sub-task of text-to-motion synthesis, we train our method on HumanML3D~\cite{humanml3d} dataset. 
It consists of ${\approx}28k$ text-annotated motion sequences from AMASS dataset~\cite{amass}.
Each sequence in the dataset is on average $7.1s$ long and has been annotated 3-4 times, thus providing a rich corpus of textual annotation for motion data.
We also use the BABEL dataset~\cite{babel} for qualitative evaluation that contains shorter phrase-level motion annotations. 
\begin{figure*}[!th]
    \includegraphics[width=\textwidth ]{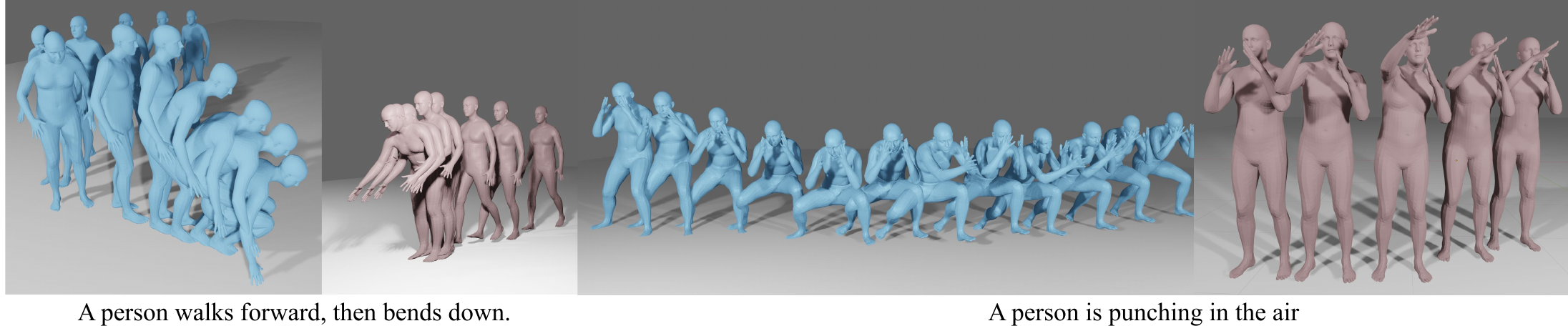} 
    \captionof{figure}{\textbf{Examples of diverse motion generation for a given text prompt.}  Notice the variations in terms of the \textit{direction} of movement as well as the difference in stances. More results, especially for choreography synthesis, can be found in the supplementary video.
    }
    \label{fig:visualisations} 
\end{figure*}
 
\noindent \textbf{Evaluation Metrics:}
Similar to dance synthesis, we evaluate the synthesised motions using the conventionally used evaluation metrics on HumanML3D dataset: Average Pairwise Euclidean Distance  (Diversity) and Multi-Modality. 
The \textit{multi-modality} metric evaluates the per-prompt diversity claims of the method by sampling the method $N$ times for the same text input and computing the average pairwise Euclidean distance of the synthesised motions; a higher Euclidean distance signifies higher variations.
Similarly, \textit{diversity} metric computes the average pairwise Euclidean distance between random pairs in the dataset, irrespective of the input prompt.
Finally, the R-Precision score measures the classification accuracy of the synthesised motions on a pre-trained classifier~\cite{humanml3d}. 
However, our network represents motion using joint positions, whereas the classifier network requires an over-parameterised representation of motion involving joint positions, 6D joint angles, local velocities and root translation. 
%
To make our method compatible for evaluation, we derive the remaining inputs based on the joint positions using inverse kinematics on the estimated joints. 
We provide qualitative results in Fig.~\ref{fig:teaser} of music-to-dance and text-to-motion synthesis results as well as in the supplementary video. 
%
\par
\noindent \textbf{Quantitative Results:} 
Table~\ref{tab:text} illustrates the performance of our method on the HumanML3D dataset. 
Similar to the case of Music-to-Dance synthesis, our method achieves state-of-the-art results in terms of synthesis variety.
This is exhibited by our performance on the multi-modality metric ($2.52$ vs $2.09$). 
Further, our diversity score of $8.82$ is similar to the ground truth $9.5$ and second only to Guo~\etal's~\cite{humanml3d} $9.18$.
Recall that to calculate the R-Precision (and FID) defined in T2M, we needed to perform IK on the synthesized joint positions.
Naturally, this is suboptimal and prohibitively prone to jittery motion (esp.~around the head joint) and manifests itself in worse R-Precision score of $0.492$ compared to $0.74$ of the state of the art method.
We now discuss the perceptual evaluation of our results through a user study.

\begin{table}[!t]
\resizebox{\columnwidth}{!}{%
\begin{tabular}{lccc}
\hline 
Methods                                  & Diversity~$\rightarrow$       & Multi-Modality$\uparrow$   & R-Precision~$\uparrow$       \\ \hline
Real Motions                            & 9.503                & n/a              & 0.797      \\ \hline
Language2Pose~\cite{jl2p}               & 7.676                & n/a              & 0.486      \\
Text2Gesture~\cite{bhattacharya2021text2gestures}        & 6.409                & n/a              & 0.345      \\
MoCoGAN~\cite{Tulyakov:2018:MoCoGAN}                  & 0.462                & 0.019          & 0.106      \\
Dance2Music~\cite{li_dance}             & 0.725                & 0.043          & 0.097      \\
Guo~\etal~\cite{humanml3d}              & \textbf{9.188}                & {2.090}          & \textbf{0.740}      \\ \hline
MoFusion (\textit{Ours})                & \textit{\textbf{8.82}}                 & \textbf{2.521}          & \textit{\textbf{0.492}}      \\ \hline
\end{tabular}
}
\caption{Comparison of our method with the previous state of the art on HumanML3D.} 
\label{tab:text}
\end{table}

\begin{figure}[!t]
    \centering
    \includegraphics[width=\linewidth ]{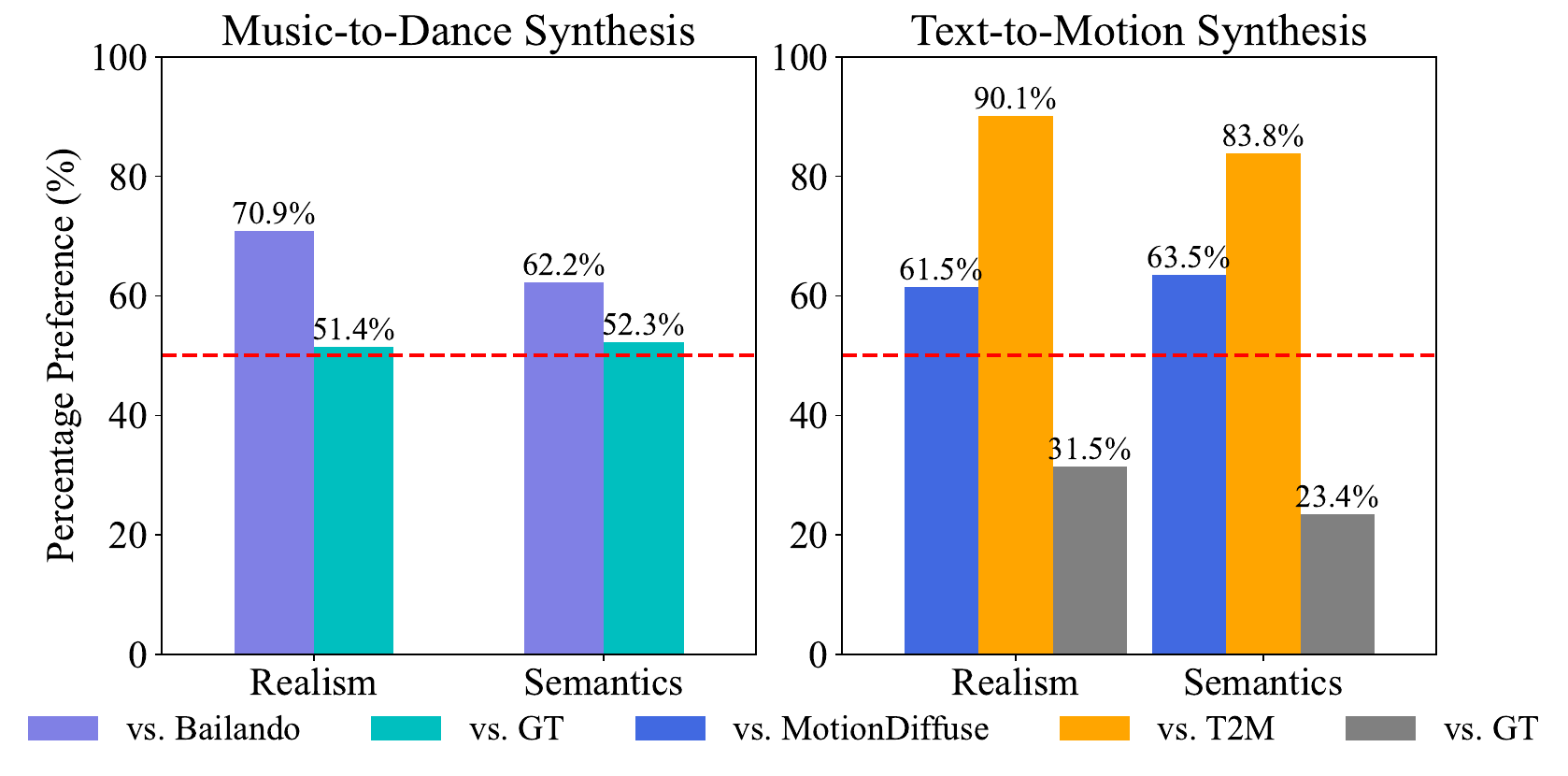}
    \captionof{figure}{The results of the user study are based on metrics of \textit{Realism} and \textit{Semantics}. As explained by Sec.~\ref{sec:user_study}, realism measures \textit{how realistic is the motion shown as a prompt} and semantics measures \textit{how well a motion corresponds to music/text}. Each bar indicates the user preference for motion generated by MoFusion compared to another motion. 
    }
    \label{fig:user_study}  
\end{figure}
\subsection{User Study} 
\label{sec:user_study}
It is worth noting that all the evaluation metrics discussed above are imperfect performance indicators.
Thus, Diversity and Multi-modality can be fooled by an untrained network that produces random, but meaningless, motion every time it is sampled.
Likewise, the FID metric used in AI Choreographer~\cite{aichoreo} uses hand-crafted features and incorrectly rewards overfitting. 
We, therefore, conduct a user study wherein we invite participants to perceptually evaluate the quality of our synthesis. 
To that end, we randomly sample audio (or text) queries from the test set and present each participant with two options to choose from. 
One of the two options shows our synthesis, and the other option can come from either the ground truth or from other state-of-the-art methods; Bailando~\cite{bailando} for audio and T2M~\cite{humanml3d} and MotionDiffuse~\cite{motiondiffuse} for text.
After having seen the two motions, the users are asked to answer the following two questions: \textit{``Which motion best justifies the music/text prompt?''} and \textit{``Which motion looks more realistic?''}.
This way, we evaluate the methods on their Semantic accuracy as well as Realism.
Fig.~\ref{fig:user_study} informs the results of the user study.
We achieve better semantic accuracy than T2M and MotionDiffuse.
It is interesting to note that our synthesis was considered more realistic than the ground-truth choreography on $51.4\%$ of occasions.
We also do well on semantics ($52.3\%$), primarily because the ground-truth choreography consists of several \textit{basic} motions in which not much dancing takes place.
More comparisons can be found in our video. 
\subsection{Interactive Motion Editing}
\label{sec:editing}
\par
\noindent \textbf{Seed-Conditioned Motion Synthesis:} In this setting, the goal is to forecast future motion frames based on a user-provided seed sequence of a few frames.
For our analysis, we consider a seed sequence of $S = 40$ frames ($2s$) and synthesise the future $N = 160$ frames. 
To achieve this, we first construct \textit{noise} vectors by forward-diffusing the seed frames to produce $P^{(t)} \in \mathbb{R}^{(S+N) \times 3J}$ for each time step $t$ wherein the remaining $N$ frames are populated with random noise.
Then, at each denoising step, we use a mask $\delta$ to ensure that the seed frames are not denoised. 
Thus, the resulting motion looks at a snapshot of the diffused seed sequence at every denoising step and generates a faithful motion corresponding to this snapshot. 
\par
\noindent \textbf{Motion Inbetweening:} In a manner similar to seed-conditioned synthesis, we perform motion inbetweening by fixing a set of keyframes in the motion sequence and reverse-diffusing  the remaining frames. 
This application is of significant utility for virtual character animation as it provides an easy way to in-fill the keyframes. 
\section{Discussion and Conclusion}\label{sec:conclusions}
\noindent\textbf{Discussion:} 
Through our analysis, we highlighted the ability of Denoising-Diffusion Probabilistic Models for conditional motion synthesis.
A less-discussed aspect of MoFusion is its ability to avoid convergence to mean pose, especially since the motion is synthesised in a non-autoregressive manner.
Thanks to a large latent space, it also avoids motion flicker artefacts that quantised codebook-based methods~\cite{tm2t} are prone to.
Finally, two aspects of our model that could be improved in future are 1) the inference time and 2) comparably restricted vocabulary for textual conditioning.  
At the same time, we foresee that MoFusion will benefit in future from fundamental advances in diffusion models and more richly annotated datasets. 

\noindent\textbf{Concluding Remarks:} 
All in all, we introduced the first approach for 3D human motion synthesis based on diffusion models. 
The proposed MoFusion method accepts audio or textual conditioning signals and produces temporally-coherent human motion sequences that are longer, more diverse and more expressive compared to the outputs of previous approaches. 
Our claims are supported by thorough experiments and a user study.
Moreover, MoFusion has direct applications in computer graphics, such as virtual character animation and crowd simulation. 
We interpret the obtained results as an encouraging step forward in  cross-modal generative synthesis in computer vision. 

\noindent\textbf{Acknowledgements.} 
This work was supported by the ERC  Consolidator Grant \textit{4DReply} (770784).

{\small
\bibliographystyle{ieee_fullname}
\bibliography{egbib}
}

\clearpage
\maketitlesupplementary

This supplementary document provides additional ablation results (Sec.~\ref{sec:results}), additional implementation details (Sec.~\ref{sec:implementation}) and finally, additional details on the user study (Sec.~\ref{sec:user_study_details}),  

\section{Additional Results}
\label{sec:results}
\noindent \textbf{Perfomance with different training objectives:}
In this work, we present different training objectives to optimize the MoFusion Architecture.
We measure the effect of different losses to gauge how well dance sequences align with music beats.
Our results in Table~\ref{tab:ablation_losses} show that as we incorporate different kinematic losses, the Beat Alignment Score improves which demonstrates better music-to-dance synthesis quality. 
\par
By addition of kinematic losses in MoFusion framework, we observe better performance than the state of the art in Beat Alignment Score (BAS) and our best BAS is even better than the ground truth data (0.237). 
This is due to better generation quality which matches music beats across the motion sequence.
Note, $\mathcal{L}_{da}$ with generation length 10 seconds (first row in our method's results) is the configuration we use to compare results with the state-of-the-art methods in Table 1 in the main draft.
However, for ablation study, we use a max. motion generation length of 20 seconds as the ablations are clearer in this setting.
\begin{table}[h]
\begin{tabular}{ll}
\hline
Method                                                         & BAS                       \\ \hline \hline
Ground Truth                                                   & 0.237                     \\ \hline
Li \etal 
& 0.160                     \\
DanceNet 
& 0.143                     \\
Dance Revolution 
& 0.195                     \\
AI Choreographer
& 0.221                     \\
Bailando
& 0.233                     \\ \hline
Ours($\mathcal{L}_{da}$) - 10 sec                                              & \multicolumn{1}{r}{0.230}  \\
Ours($\mathcal{L}_{da}$)                                                       & \multicolumn{1}{r}{0.234} \\
Ours($\mathcal{L}_{da} + \mathcal{L}_{m}$)                                                  & \multicolumn{1}{r}{0.242} \\
Ours($\mathcal{L}_{da} + \mathcal{L}_{m} + \mathcal{L}_{s} + \mathcal{L}_{a}$)                                             & \multicolumn{1}{r}{0.252} \\ \hline
\end{tabular}
\centering
\caption{Comparison between performance on BAS by training our network with different training objectives. Here, ``10 sec" refers to length of motion generation. All other models of MoFusion had a generation length of 20 seconds.}
\label{tab:ablation_losses}
\end{table}

\noindent \textbf{Music-to-Dance Synthesis with Seed Motion Input:}
Previous methods~\cite{aichoreo, li_dance, bailando}  synthesize dance motion with a seed pose as input which guides the training process and dance generation process as well.
However, we do not train our method with seed motion as input. 
Instead, we synthesize the dance motion from scratch which is solely conditioned on melspectrogram of dance music.
To test performance of our model with a seed sequence, we test the model performance by running reverse-diffusion process at test time with a \textit{seed sequence}. 
The seed sequence consists of first two seconds of ground-truth motions and we predict a motion sequence in correspondence with first two seconds of input. 
We follow~\cite{aichoreo} while choosing the length of seed sequence as 2 seconds. 
\par
As observed in Table~\ref{tab:seed_motion}, there is an overall increase in Beat Alignment Score due to addition of ground truth data and here, we again observe a trend of higher performance in BAS as we add more losses. 
It is noteworthy that we do not retrain with seed motion input. Rather, we use a pretrained dance synthesis model to perform inference with seed input.
Our supplementary video shows results for seed input synthesis wherein we can observe smooth transition from seed input sequence to forecast dance sequence.

\begin{table}[h]
\begin{tabular}{llr}
\hline
Seed Motion                 & Length (sec) & \multicolumn{1}{l}{BAS} \\ \hline \hline
$\mathcal{L}_{da}$                    & 10           & 0.257                   \\
\hline
$\mathcal{L}_{da}$                    & 20           & 0.269                   \\
$\mathcal{L}_{da} + \mathcal{L}_{m}$        & 20           & 0.264                   \\
$\mathcal{L}_{da} + \mathcal{L}_{m} + \mathcal{L}_{s} + \mathcal{L}_{a}$ & 20           & 0.265                   \\ \hline
\end{tabular}
\centering
\caption{Performance comparison of different trained networks at inference with a seed sequence. Here, Length refers to motion generation length in seconds.}
\label{tab:seed_motion}
\end{table}

\section{Implementation Details}
\label{sec:implementation}
\noindent \textbf{Diffusion Model:} 
We use 1000 diffusion steps as $T$ for the diffusion process and change the variances $\beta_t$ linearly from $0.0001$ to $0.02$. 
For training our framework, we use single NVIDIA RTX A40 with task-specific batch sizes. 

\par
\noindent \textbf{Music-to-Dance Synthesis:} 
We use a latent dimension of 1024 in the audio encoder which takes melspectrogram as input. 
In 1D-UNet model, we use cross-modal transformer blocks with 16 attention heads and a cross attention dimension of 1024. 
We employ AdamW~\cite{adamw} as an optimizer with a learning rate of $5~\times~10^{-4}$.
Moreover, we use a batch size of 32 for optimization.
\par
To represent motion, we use 24 joint positions from SMPL model data which we extract from AIST++ Dataset~\cite{aichoreo}. 
We opt to train our model on 3D joint positions as joint angle representation performed worse during our experiments.
As we observed from our experiments, our framework can also be trained with other joint position representations like COCO Keypoints format~\cite{cocodataset}.

\par
\noindent \textbf{Text-to-Motion Synthesis:}
For text encoder, we use CLIP ViTB/32~\cite{clip} with a latent dimension of $512$ for text encoding. 
%
The batch-size used is 128 and the network is trained with AdamW optimizer and a learning rate of $0.0002$ is used.
Following the conventional diffusion wisdom, we use a warm-up schedule of 500 iterations in the beginning.
As discussed in the main draft, we use 22 joint positions of SMPL-X model data to represent motion. %
This is extracted from SMPL data given in HumanML3D dataset~\cite{humanml3d}.

\par

\begin{figure}
    \centering
    \includegraphics[width=\linewidth]{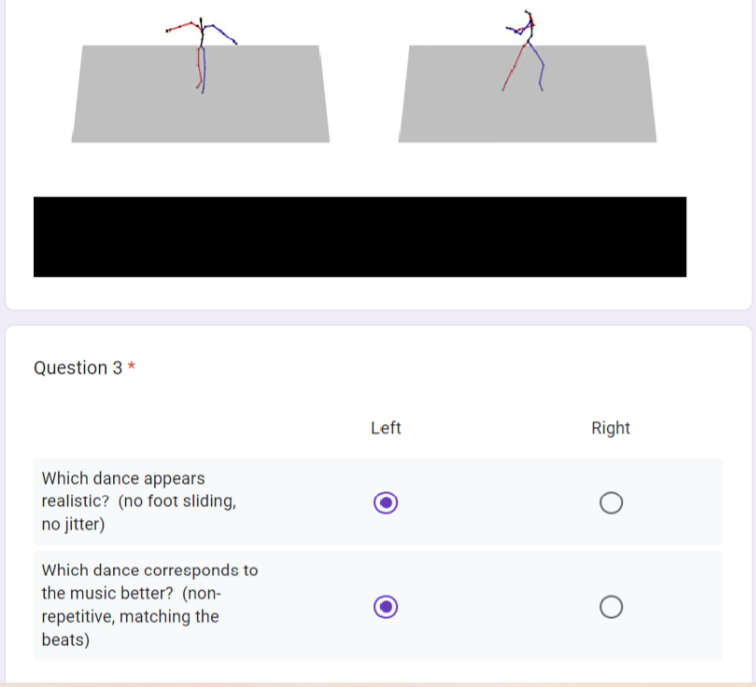}
    \caption{A screenshot of the user study. The participants were asked 18 such questions.}
    \label{fig:user_study}
\end{figure}

\section{Details of User Study}\label{sec:user_study_details} 
We conducted the user study with $40$ participants, with each participant answering $18$ questions which asked the users to compare our synthesis results with other state-of-the-art methods.
The participants took 8-10 minutes to submit their responses.
We provide a snapshot of the interface in Fig.~\ref{fig:user_study}

\end{document}